\pdfoutput=1

\documentclass[11pt]{article}

\usepackage[preprint]{acl}
\usepackage{tabularx}
\usepackage{booktabs}  
\usepackage{siunitx} 
\usepackage{times}
\usepackage{latexsym}

\usepackage[T1]{fontenc}

\usepackage[utf8]{inputenc}

\usepackage{microtype}

\usepackage{inconsolata}

\usepackage{graphicx}
\usepackage{relsize}

\usepackage{lmodern}
\usepackage{listings} 
\usepackage{hyperref}

\lstdefinelanguage{custom}{
  basicstyle=\ttfamily\small, 
  breaklines=true, 
  breakatwhitespace=true, 
  columns=fullflexible, 
}

\title{CRISP: Complex Reasoning with Interpretable Step-based Plans}

\author{
  Matan Vetzler\thanks{Matan Vetzler and Koren Lazar contributed equally to this work.}\\
  IBM Research, Israel\\
  \texttt{matan.vetzler@ibm.com} \\
  \And
  Koren Lazar\footnotemark[1]\\
  IBM Research, Israel\\
  \texttt{koren.lazar@ibm.com}
  \And
  Guy Uziel\\
  IBM Research, Israel\\
  \texttt{guy.uziel1@ibm.com}
  \AND
  Eran Hirsch\\
  IBM Research, Israel\\
  \texttt{eran.hirsch@ibm.com}
  \And
  Ateret Anaby-Tavor\\
  IBM Research, Israel\\
  \texttt{atereta@il.ibm.com}
  \And
  Leshem Choshen\\
  IBM Research, MIT\\
  \texttt{leshem.choshen@ibm.com}
}

\begin{document}
\maketitle
\begin{abstract}  
Recent advancements in large language models (LLMs) underscore the need for stronger reasoning capabilities to solve complex problems effectively. While Chain-of-Thought (CoT) reasoning has been a step forward, it remains insufficient for many domains. A promising alternative is explicit high-level plan generation, but existing approaches largely assume that LLMs can produce effective plans through few-shot prompting alone, without additional training. In this work, we challenge this assumption and introduce CRISP (Complex Reasoning with Interpretable Step-based Plans), a multi-domain dataset of high-level plans for mathematical reasoning and code generation. The plans in CRISP are automatically generated and rigorously validated--both intrinsically, using an LLM as a judge, and extrinsically, by evaluating their impact on downstream task performance. We demonstrate that fine-tuning a small model on CRISP enables it to generate higher-quality plans than much larger models using few-shot prompting, while significantly outperforming Chain-of-Thought reasoning. Furthermore, our out-of-domain evaluation reveals that fine-tuning on one domain improves plan generation in the other, highlighting the generalizability of learned planning capabilities.


\end{abstract}
\section{Introduction}
\label{sec:introduction}
\begin{figure*}[t]
    \centering
    \includegraphics[width=\textwidth]{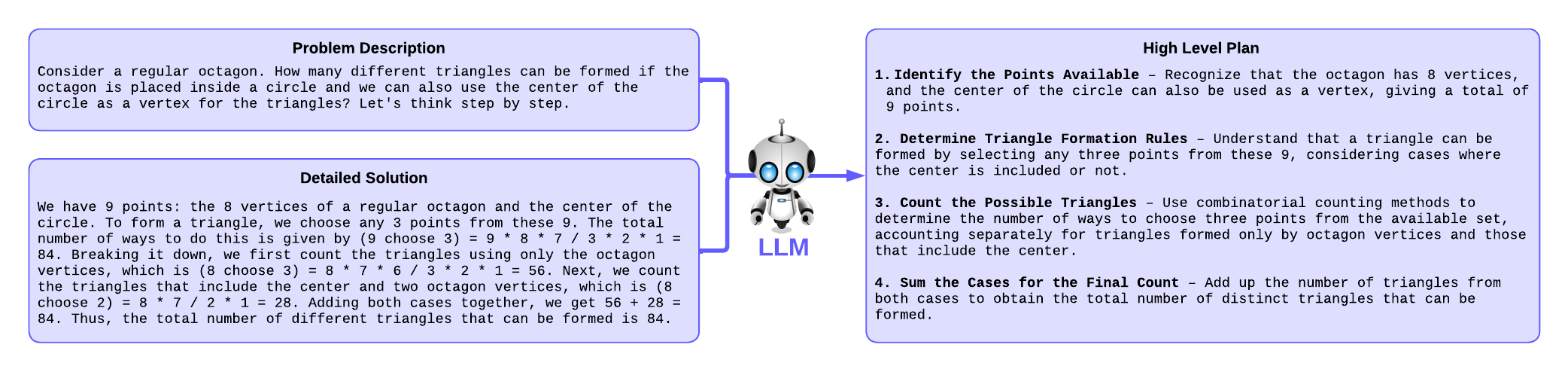} 
    \caption{Example from the Math domain showing a problem statement, its detailed solution, and the generated high-level plan. The LLM was prompted to retain the high-level strategy while omitting lower-level details. It was also provided with a few examples and guidelines tailored to the specific domain.}
    \label{fig:plan-generation-example}
\end{figure*}
    
Large language models (LLMs) abilities advance rapidly in logical reasoning, code generation, and mathematical problem-solving~\citep{plaat2024reasoninglargelanguagemodels, jiang2024surveylargelanguagemodels, ahn-etal-2024-large}.
A key factor behind recent breakthroughs is the ability of LLMs to break down complex tasks into manageable steps—an approach exemplified by chain-of-thought prompting~\citep{wei2022chain}.

While chain-of-thought reasoning has led to notable performance gains, it remains prone to errors such as missing intermediate steps and semantic misunderstandings~\citep{wei2022chain, jiang2024self}. 
To address these challenges, recent studies have explored prompting strategies that explicitly break down problems into subtasks~\citep{dua2022successive, zhou2023leasttomost, khot2023decomposed, prasad-etal-2024-adapt, ding2024semcoder}. 
One prominent approach is self-generating a high-level plan. by the LLM before executing the task. This plan-and-solve approach lead to significant improvements in mathematical, commonsense, and symbolic reasoning, as well as code generation~\citep{wang-etal-2023-plan, jiang2024self}. 
However, the self-generated plans were only partially effective, as they did not match the performance of ground-truth planning across various downstream tasks.

In this work, we argue that generating high-quality high-level plans is a hard challenge for LLMs, as current models often struggle to produce accurate and effective plans across different domains. 
Part of this difficulty arises from the scarcity of explicit planning data, as people rarely externalize or document their high-level reasoning in a structured way.
To address this gap, we introduce CRISP (Complex Reasoning with Interpretable Step-based Plans), a novel dataset designed to enhance high-level planning capabilities.
CRISP spans two domains: mathematics and code generation—where solutions naturally decompose into structured, high-level steps. 
These plans were derived from annotated detailed solutions of Magpie-Reasoning-V1-150K~\citep{xu2025magpie} and validated both extrinsically, by measuring their impact on the original task performance, and intrinsically, using LLM-based judgment to assess coherence, conciseness, clarity, and completeness.

To assess CRISP's impact, we perform a series of experiments on four reasoning-related datasets: MBPP~\citep{austin2021program}, HumanEval~\citep{chen2021evaluating}, GSM8K~\citep{cobbe2021training}, and MATH~\citep{hendrycks2021measuring}. Our findings indicate that while larger models generate plans that lead to better performance on the reasoning-related benchmarks, a small model can surpass them with a lightweight fine-tuning on CRISP using LoRA~\citep{hu2022lora}.
Furthermore, with lightweight fine-tuning with LoRA~\citep{hu2022lora} on our dataset, LLMs exhibit substantial improvements in plan generation, as reflected in both higher performance on structured reasoning benchmarks such as MBPP~\citep{austin2021program}, HumanEval~\citep{chen2021evaluating}, and higher quality scores assigned by LLM-based evaluations.

Additionally, we assess the out-of-domain generalization of planning abilities and find that CRISP effectively transfers these capabilities across different tasks and domains. For example, a model fine-tuned on the Math domain achieves a pass@1 score of 84.6 on the HumanEval code dataset—only 0.4 points lower than the same model fine-tuned on the Coding and Debugging domain.
This strong transferability highlights CRISP's potential for seamless integration into existing training pipelines, where it can enhance LLM reasoning abilities and improve performance across diverse downstream tasks.

The significance of this work lies in its emphasis on high-level planning as a beneficial trainable capability that current off-the-shelf models do not excel in. 
By providing LLMs with explicit fine-tuning on high-level planning, we enhance their ability to decompose tasks which in turn improves their applicability to real-world scenarios requiring robust, domain-agnostic reasoning.
The dataset is publicly available \href{https://huggingface.co/collections/MatVet/crisp-high-level-plans-67b1c2a02cfc42b4bc8a0271}{here}.

Our contributions are threefold:
\begin{itemize}
    \item We show that generating high-level plans is a \textbf{challenging task} for LLMs, yet highly beneficial. 
    \item We introduce the \textbf{CRISP dataset}, a multi-domain dataset of high-level plans derived from annotated detailed solutions.
    \item We demonstrate that even a short LoRA-based \textbf{fine-tuning improves} the quality of generated plans, significantly outperforming larger models.
    \item We show that high-level \textbf{planning abilities transfer well} between tasks and domains through out-of-domain evaluation.
\end{itemize}

\section{Related Work}
\begin{figure*}[t]
    \centering
    \includegraphics[width=\textwidth]{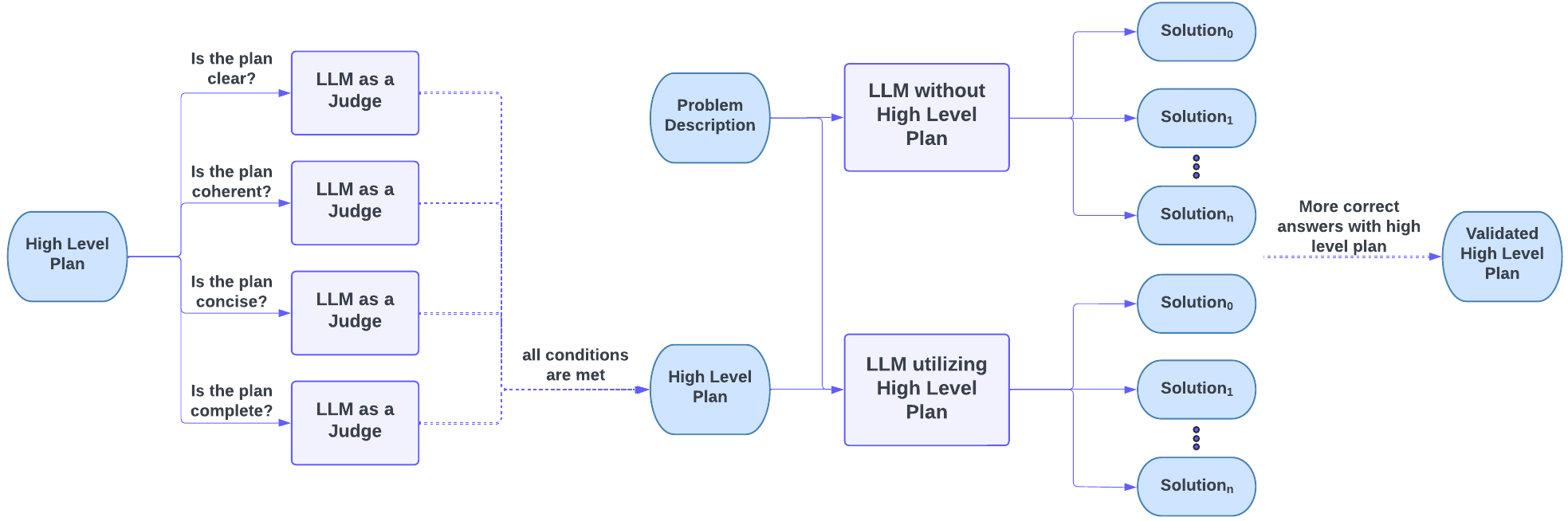} 
    \caption{The validation and filtering pipeline of CRISP. Each generated high-level plan undergoes binary validation, where its clarity, coherence, conciseness, and completeness are assessed by the LLM. If all attributes receive a positive judgment, the plan is then externally validated by comparing the solutions generated with and without the high-level plan in the prompt, filtering out those that do not lead to improved performance on the original task.
    }
    \label{fig:plans-validation-pipeline}
\end{figure*}

A growing body of research investigates methods to enhance LLMs' performance on tasks requiring multi-step reasoning and structured problem-solving.
One prominent approach is Chain of Thought (CoT) reasoning~\citep{NEURIPS2022_9d560961}, which improves LLMs' ability to handle complex tasks by explicitly breaking down reasoning into intermediate steps. 
This method has significantly improved performance on arithmetic, commonsense, and symbolic reasoning benchmarks.

Building on CoT, several works have introduced further improvements, such as exploring multiple reasoning trajectories~\citep{wang2023selfconsistency}, backtracking and applying search algorithms~\citep{yao2023tree, hao-etal-2023-reasoning, Besta_2024, zhou2024language}, self-evaluation~\citep{xie2023selfevaluation}, self-reflection~\citep{shinn2023reflexion}, and self-refinement~\citep{madaan2023selfrefine}. These methods enable models to explore multiple solution paths, iteratively improving their responses.

Another line of work focuses on explicit task decomposition, where problems are broken down into structured subtasks. 
Some approaches generate successive questions\citep{dua2022successive, zhou2023leasttomost}, others decompose problems into lines of code\citep{chen2023program, yang2023intercode, ding2024semcoder}, and some define explicit hierarchical subtasks~\citep{khot2023decomposed}.
Most similar to our work are "Plan-and-Execute" approaches, where an LLM first generates a structured plan before executing it to solve a problem~\citep{wang-etal-2023-plan, jiang2024self}. 
Additionally, \citet{prasad-etal-2024-adapt} proposed iteratively refining plans upon execution failures. 
However, these methods have largely been evaluated in single-domain settings, and they treat high-quality plan generation as an emergent ability, requiring no additional training. 
In contrast, our work systematically explores plan generation across multiple domains, demonstrating both the challenges of this task and the tangible benefits of training LLMs to produce structured plans.
However, these approaches are primarily benchmarked within a single domain and consider the task of generating a high-quality plan as an emergent capability requiring no additional training. 
In contrast, our work systematically explores plan generation across multiple domains, demonstrating both the challenges of this task and the benefits of training LLMs to generate structured plans.

Several datasets incorporate task decomposition, including those designed for web agents\citep{zheran2018reinforcement, NEURIPS2022_82ad13ec, deng2023mindweb, pmlr-v70-shi17a}, household activities\citep{puig2018virtualhome, shridhar2021alfworld, shridhar2020alfred}, games\citep{guss2019minerl, prasad-etal-2024-adapt}, and robotics\citep{10802322, zhang2023bootstrap, li2023interactive}. 
These datasets primarily focus on dynamic decision-making in interactive environments, where planning is contingent on real-time feedback and reinforcement learning. In contrast, CRISP is designed for structured, multi-domain task decomposition, emphasizing myopic problems—tasks that can be solved through a predefined sequence of steps rather than adaptive decision-making. 

\section{Dataset Collection}
\label{sec:dataset}
In this section we elaborate on the creation of CRISP. In Section~\ref{subsec:plan-generation} we describe how we generated the high-level plans based on careful prompt engineering. 
Then, in Section~\ref{subsec:plan-filtering} we describe the validation and filtering mechanisms we developed to ensure the quality of the generated high-level plans based on both intrinsic and extrinsic evaluations. Finally, in Section~\ref{subsec:dataset-analysis}, we analyze how the filtering and validation affect downstream tasks and the scores assigned to the plans based on LLM-based judgment.

\subsection{High-Level Plan Generation}
\label{subsec:plan-generation}

Our high-level plan dataset is derived from Magpie-Reasoning-V1-150K~\citep{xu2025magpie}, which is licensed under `Llama3' and spans two domains: \emph{Math} and \emph{Coding and Debugging}.
Magpie reasoning examples in the domains of math and coding and debugging illustrate how models can decompose complex problems into structured subproblems. In math, this involves breaking down equations or proofs into intermediate steps, while in coding and debugging, it includes identifying error patterns, generating hypotheses about potential bugs, and testing fixes iteratively.
Each instance includes a problem statement and a detailed solution, generated by \href{https://huggingface.co/Qwen/Qwen2-72B-Instruct}{Qwen2-72B-Instruct} for math and \href{https://huggingface.co/meta-llama/Meta-Llama-3-70B-Instruct}{Llama-3-70B-Instruct} for coding. An example of such a problem statement and its corresponding detailed solution is depicted on the left side of Figure~\ref{fig:plan-generation-example}.
The dataset encompasses a wide range of topics from mathematics and coding such as geometry, algebra, and integrals, differential equations, and probability in mathematics, as well as data structures, algorithm design, syntax and logic error fixes, concurrency, and general software engineering in coding and debugging. 
In total, we extract 74,225 math-related samples and 66,342 coding-related samples.
Since the problems in this dataset are myopic—meaning they can often be solved using a predefined sequence of steps—we believe that generating a high-level plan should be particularly beneficial.

For each problem and its detailed solution, we used \href{https://huggingface.co/mistralai/Mixtral-8x7B-Instruct-v0.1}{Mixtral-8x22B-Instruct-v0.1} to generate a high-level plan through few-shot prompting. We selected this model after manually evaluating its generated plans and finding them to be of higher quality than those from other LLMs, along with its permissive license.
An example of an such a high-level plans is depicted on the right side of Figure~\ref{fig:plan-generation-example}.
In the plan generation prompt, we instructed the model to outline the high-level logical strategy while abstracting away implementation details. This approach ensures that information from the detailed solution, which the model should not have prior knowledge of, remains undisclosed. Simultaneously, it preserves a degree of flexibility, allowing the model to determine the precise method for executing each step at a later stage.
The full generation prompt of Math is provided in Appendix~\ref{subsec:appendix-prompt-for-plan-generation-in-crisp}.




\subsection{Filtering and Validation}
\label{subsec:plan-filtering}

After gathering a substantial collection of high-level plans, we implemented a two-step filtering process to validate the plans intrinsically and extrinsically, as illustrated in Figure~\ref{fig:plans-validation-pipeline}. 
First, we apply `LLM as a Judge' with \href{https://huggingface.co/meta-llama/Llama-3.1-70B-Instruct}{Llama-3.1-70B-Instruct} to determine whether the generated plans are \textbf{concise, clear, coherent, and complete}. These four attributes are essential for ensuring that a plan is described efficiently without redundancy or repetition (concise), is easily understandable without ambiguity or vague language (clear), follows a logical sequence without missing critical transitions (coherent), and includes all the essential steps to fully address the problem and derive the solution (complete). 
We believe that ensuring these four attributes is crucial for generating high-quality plans that are both interpretable and actionable, facilitating their usefulness in various downstream tasks.
Each attribute is assessed using a binary judgment, and any plan that fails to meet one or more criteria is discarded. In this step 7,412 math plans and 5,592 code plans were filtered out, accounting for approximately 9\% of the original dataset. For the full prompt used in this evaluation, refer to Appendix~\ref{subsec:appendix-llm-as-a-judge-prompt}.

After the intrinsic validation, we assessed the generated plans by testing their impact on the model’s ability to successfully complete the original tasks.
To that end, we generated 10 solutions both with and without the plan using Llama-3.1-70B-Instruct and discarded cases where the number of correct final answers was higher without the plan.
This step removes an additional 1,089 math plans and 4,612 code plans that---while clear, concise, coherent, and complete---failed to produce more correct answers than when the LLM was not provided with a high-level plan.

After filtering, we retain 65,800 math plans and 56,200 coding plans. Table~\ref{tab:dataset-stats} summarizes the final dataset. 




\subsection{Dataset Analysis}
\label{subsec:dataset-analysis}
Empirical validation shows that despite reducing the dataset size, each filtering step improves quality enough to warrant the deletion. Specifically, it improves performance on external benchmarks: The intrinsic filtering stage increases accuracy by 0.72 points on GSM8K~\citep{cobbe2021training} and MATH~\citep{hendrycks2021measuring} and 0.44 points on MBPP~\citep{austin2021program} and HumanEval~\citep{chen2021evaluating}. 
Similarly, the extrinsic filtering, which validates the impact on the original task, further improves accuracy by 0.28 points for mathematics and 0.32 points for coding.
These improvements confirm that our filtering pipeline successfully distills the datasets into high-quality high-level plans that are concise, clear, coherent, and complete, as well as beneficial on downstream tasks.


\begin{table}[t]
\begin{center}
\begin{sc}
\resizebox{\columnwidth}{!}{
\begin{tabular}{@{}lll@{}}
\toprule
Domain          & \# Examples & \# Steps \\ \midrule
Math            & 65,800      & 3.8      \\
Code\&Debugging & 56,200      & 4.4      \\ \bottomrule
\end{tabular}
}
\caption{Statistics by domain in CRISP after applying filtering. We report the total number of generated instances, and the average number of steps per plan.}
\label{tab:dataset-stats}
\end{sc}
\end{center}
\end{table}

\section{Experiments}
\label{sec:experiments}

\begin{table*}[t]
\centering
\begin{sc}
\resizebox{\textwidth}{!}{
\begin{tabular}{llcccc}
\toprule
\textbf{Planner} & \textbf{Solver} & \textbf{MBPP} & \textbf{HumanEval} & \textbf{GSM8K} & \textbf{MATH} \\
& & \textbf{Pass@1\textsmaller{(Err $\downarrow$)}} & \textbf{Pass@1\textsmaller{(Err $\downarrow$)}} & \textbf{Acc.\textsmaller{(Err $\downarrow$)}} & \textbf{Acc.\textsmaller{(Err $\downarrow$)}} \\
\midrule
Standard ICL (No Plan) & Small & 53.2 & 66.8 & 75.1 & 41.7 \\
CoT (No Plan) & Small & 60.1 & 71.2 & 84.6 & 49.3 \\
Vanilla Small & Small & 60.6 \textsmaller{(\textcolor{ForestGreen}{1.3\%$\downarrow$})} & 71.9 \textsmaller{(\textcolor{ForestGreen}{2.4\%$\downarrow$})} & 84.9 \textsmaller{(\textcolor{ForestGreen}{1.9\%$\downarrow$})} & 49.6 \textsmaller{(\textcolor{ForestGreen}{0.6\%$\downarrow$})} \\
Vanilla Large & Small & 62.0 \textsmaller{(\textcolor{ForestGreen}{4.98\%$\downarrow$})} & 73.1 \textsmaller{(\textcolor{ForestGreen}{6.6\%$\downarrow$})} & 85.7 \textsmaller{(\textcolor{ForestGreen}{7.1\%$\downarrow$})} & 53.2 \textsmaller{(\textcolor{ForestGreen}{7.7\%$\downarrow$})} \\
Fine-tuned Small & Small & \textbf{64.8 \textsmaller{(\textcolor{ForestGreen}{11.8\%$\downarrow$})}} & \textbf{76.4 \textsmaller{(\textcolor{ForestGreen}{18.1\%$\downarrow$})}} & \textbf{87.1 \textsmaller{(\textcolor{ForestGreen}{16.2\%$\downarrow$})}} & \textbf{60.6 \textsmaller{(\textcolor{ForestGreen}{22.3\%$\downarrow$})}} \\
\midrule
Standard ICL (No Plan) & Large & 69.6 & 76.1 & 88.6 & 58.4 \\
CoT (No Plan) & Large & 73.7  & 80.8  & 94.3  & 67.2  \\
Vanilla Small & Large & 73.4 \textsmaller{(\textcolor{red}{1.1\%$\uparrow$})} & 80.3 \textsmaller{(\textcolor{red}{2.6\%$\uparrow$})} & 93.6 \textsmaller{(\textcolor{red}{1.2\%$\uparrow$})} & 65.8 \textsmaller{(\textcolor{red}{4.3\%$\uparrow$})} \\
Vanilla Large & Large & 74.1 \textsmaller{(\textcolor{ForestGreen}{1.5\%$\downarrow$})} & 82.2 \textsmaller{(\textcolor{ForestGreen}{7.3\%$\downarrow$})} & 94.8 \textsmaller{(\textcolor{ForestGreen}{8.8\%$\downarrow$})} & 70.2 \textsmaller{(\textcolor{ForestGreen}{9.1\%$\downarrow$})} \\
Fine-tuned Small & Large & \textbf{76.2 \textsmaller{(\textcolor{ForestGreen}{9.5\%$\downarrow$})}} & \textbf{85.3 \textsmaller{(\textcolor{ForestGreen}{23.4\%$\downarrow$})}} & \textbf{95.9 \textsmaller{(\textcolor{ForestGreen}{28.1\%$\downarrow$})}} & \textbf{73.1 \textsmaller{(\textcolor{ForestGreen}{18.0\%$\downarrow$})}} \\
\bottomrule
\end{tabular}
}
\caption{Evaluation on code generation and math benchmarks across plan generator models and solution generator models. `Err$\downarrow$' represents the relative reduction in error compared to the baseline of Chain-of-Thought prompting (`CoT (No Plan)')  with the same solver. `Small' refers to Granite-3.1-8B-Instruct model and `Large' refers to Llama-3.1-70B-Instruct model. Notably, the fine-tuned Granite gains the largest improvement in results across the four domains and solver models.}
\label{tab:merged-results}
\end{sc}
\end{table*}

\subsection{Plan Generator Training}
\label{subsec:plan-generator-training}

To demonstrate the practical benefits of CRISP, we applied LoRA parameter-efficient fine-tuning \citep{hu2022lora} on a small model to show that effective high-level plan generation is a learned capability; consequently, even a relatively small model, when efficiently fine-tuned on CRISP, can outperform a vanilla larger model on this task.
Specifically, we used \href{https://huggingface.co/ibm-granite/granite-3.1-8b-instruct}{Granite-3.1-70B-Instruct} model~\citep{granite24paper}.
We trained the model for five epochs with a learning rate of 1e-5 on four A100-80GB GPUs, using hyperparameters optimized through an extensive sweep. 
Additional technical details of the training procedure are provided in Appendix~\ref{appendix:lora-finetuning}. 




\subsection{Experimental Setup}
\label{subsec:experiment-setup}

To systematically assess the impact of high-level plan generation on downstream tasks, we examine various scenarios using both a small model (\href{https://huggingface.co/ibm-granite/granite-3.1-8b-instruct}{Granite-3.1-8B-Instruct} \citealp{granite24paper}) and a large model (\href{https://huggingface.co/meta-llama/Llama-3.1-70B-Instruct}{Llama-3.1-70B-Instruct}). We also experimented with another small model, Llama-3.1-8B-Instruct, and found comparable results to Granite-8B as described in Appendix~\ref{subsec:appendix-results-with-llama-8b}.
We conducted evaluations on four well-established benchmarks: the HumanEval benchmark~\citep{chen2021evaluating}, which assesses code synthesis and problem-solving capabilities, MBPP~\citep{austin2021program} which consists of around 1,000 crowd-sourced Python programming problems, GSM8K~\citep{cobbe2021training}--a grade school math word problems created by human problem writers, and the MATH benchmark~\citep{hendrycks2021measuring}, which measures performance on complex mathematical problem-solving tasks.
First, we established a baseline where the plan is based on classic Chain-of-Thought (CoT) prompting that relies solely on the problem description, which could be considered as having an emergent plan. Next, we incorporate high-level plans with plan-and-solve approach~\citep{wang-etal-2023-plan}---these plans are generated by both the small and large models via few-shot prompting without any additional training, as well as by a fine-tuned version of the small model on CRISP detailed in Section~\ref{subsec:plan-generator-training}. Once the plans are generated, another model takes them as input, along with the task description, and attempts to solve the task. We refer to the plan generation model as the `planner' and to the subsequent model as the `solver'.
We evaluated the generated plans using both the small and large models as solvers in a zero-shot setting without additional fine-tuning. 
This design enables us to directly compare the contribution of high-level plans produced by different models on various model sizes and domains.

\subsection{Extrinsic Evaluation}
\label{subsec:extrinsic-evaluation}

Table~\ref{tab:merged-results} compares the results on different planners and solvers as well as the class Chain-of-Thought (CoT), and provides valuable insights into the impact of high-level plan generation across both coding and mathematical problem-solving domains.

\paragraph{Fine-tuned planning model achieves best results.}
The plans generated by the fine-tuned small model vastly outperforms the plans generated by the vanilla models and CoT across solvers and datasets.
For example, improvements reach up to 28\% error reduction against CoT in GSM8K.
This demonstrates that fine-tuning a model for plan generation significantly benefits various myopic downstream tasks, such as code generation and mathematical problem-solving. It also shows that the plan generation capabilities of vanilla models, including larger ones like Llama-3.1-70B-Instruct, can be substantially improved, resulting in enhanced reasoning abilities.

\paragraph{Planning is better than CoT, yet the quality of plans matters.}
Plans generated by the vanilla models mostly outperformed CoT, with improvement of up to 9.1\%. Yet, while the plans generated by the large model achieved significant improvements, the plan generated by the small model achieved only minor improvement and even degradation of up to 4.3\% when using the large model as solver.
This shows that explicitly generating the high-level plans before the solution is a better approach than CoT in myopic tasks, although the quality of the plan plays a critical role.

\paragraph{The fine-tuned planner equally improves both solvers.}
Incorporating plans generated by the small fine-tuned model into the prompts of both small and large solvers results in an average error reduction of 17.1\% and 19.65\%, respectively. This indicates that both models experience similar and significant improvements from receiving a high-quality plan before generating a solution.


\paragraph{Both the planner and the solver impact performance}.
Improving either the planner or the solver leads to performance gains. However, the solver has a greater influence on overall performance. This is evident when comparing the results of a fine-tuned small planner paired with a small solver to those of a vanilla small planner paired with a large solver, where the latter configuration yields significantly better results.

\subsection{Intrinsic Evaluation}
\label{subsec:intrinsic-evaluation}

\begin{figure*}[t]
    \centering
    \includegraphics[width=\textwidth]{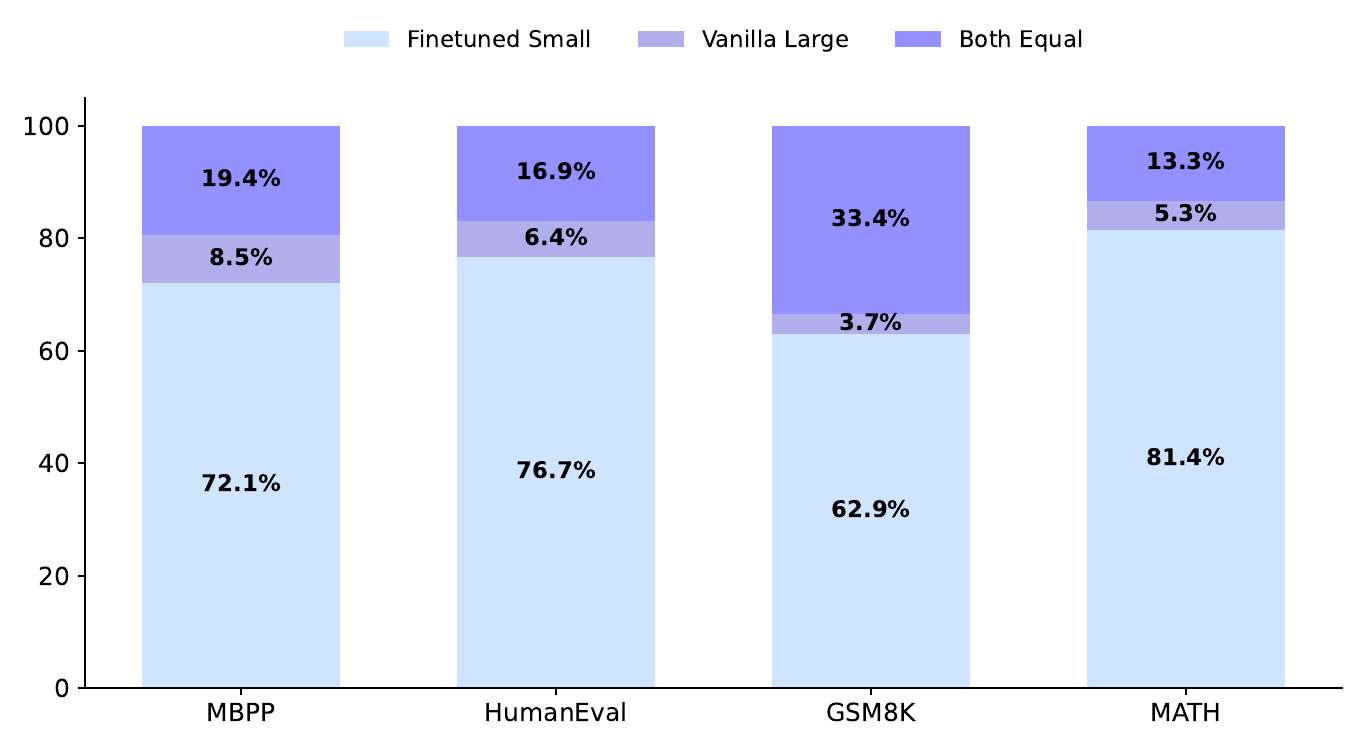} 
    \caption{LLM-based judgement comparison for clarity, coherence, conciseness, and completeness between  Granite-3.1-8B-Instruct fine-tuned on CRISP and Llama-3.1-70B-Instruct, which was also the judge. Each bar is divided into three sections representing the percentage of cases where the judge preferred the plan generated by one of the models or found both plans to be equally good.
    }
    \label{fig:llm-as-a-judge-bar-chart}
\end{figure*}

Following the extrinsic evaluation in Section~\ref{subsec:extrinsic-evaluation}, we conducted a direct comparison of plan quality using LLM-based judgment. 
Specifically, we evaluated the coherence, clarity, conciseness, and completeness of plans generated by fine-tuned Granite-3.1-8B-Instruct (described in Section~\ref{subsec:plan-generator-training}) and Llama-3.1-8B-Instruct.
We chose these two models as we would to further explore how much the fine-tuning helped compared to the best baseline across datasets. 

The results are depicted in Figure~\ref{fig:llm-as-a-judge-bar-chart}. Surprisingly, although we used Llama-3.1-70B-Instruct as both a judge and a competitor, which should create a bias toward its own generations~\citep{bitton2023visit,koo2023benchmarking}, it preferred the plans generated by the small fine-tuned model across all datasets in 73.3\% of the cases on average. Interestingly, fine-tuned Granite achieved the highest scores in the two harder datasets--MATH and HumanEval. This may indicate that the fine-tuning especially helped with plans that require more complex reasoning. The impact of our fine-tuned model's high-level plans could be speculated to stem from longer and more robust steps compared to the vanilla model's high-level plans. However, when analyzing the average number of steps in the high-level plans across the four benchmarks, we observe that the large vanilla model generates, on average, 1.3 more steps than our small fine-tuned model on coding benchmarks and 0.8 more steps on math benchmarks. This disproves such speculation and suggests that fewer, more concise, and well-structured steps have a greater impact on the final solution.

\begin{table*}[t]
\begin{center}
\begin{sc}
\begin{tabular}{@{}lllll@{}}
\toprule
Planner            & MBPP & HumanEval & GSM8K & MATH \\ 
& \textbf{Pass@1\textsmaller{(Err $\downarrow$)}} & \textbf{Pass@1\textsmaller{(Err $\downarrow$)}} & \textbf{Acc.\textsmaller{(Err $\downarrow$)}} & \textbf{Acc.\textsmaller{(Err $\downarrow$)}} \\ \midrule
CoT                & 73.7          & 80.8          & 94.3          & 67.2          \\
Vanilla Small      & 73.4 \textcolor{red}{(\textsmaller{$\uparrow$1.1\%})} & 80.3 \textcolor{red}{(\textsmaller{$\uparrow$1.5\%})} & 93.6 \textcolor{red}{(\textsmaller{$\uparrow$9.1\%})} & 65.8 \textcolor{red}{(\textsmaller{$\uparrow$6.5\%})} \\
Vanilla Large      & 74.1 \textcolor{ForestGreen}{(\textsmaller{$\downarrow$3.5\%})} & 82.2 \textcolor{ForestGreen}{(\textsmaller{$\downarrow$7.3\%})} & 94.8 \textcolor{ForestGreen}{(\textsmaller{$\downarrow$8.8\%})} & 70.2 \textcolor{ForestGreen}{(\textsmaller{$\downarrow$9.1\%})} \\
Trained Small Code & \underline{76.1} \textcolor{ForestGreen}{(\textsmaller{$\downarrow$15.4\%})} & \underline{85.0} \textcolor{ForestGreen}{(\textsmaller{$\downarrow$21.9\%})} & \textbf{95.2} \textcolor{ForestGreen}{(\textsmaller{$\downarrow$15.8\%})} & \textbf{71.9} \textcolor{ForestGreen}{(\textsmaller{$\downarrow$14.3\%})} \\
Trained Small Math & \textbf{75.4} \textcolor{ForestGreen}{(\textsmaller{$\downarrow$11.9\%})} & \textbf{84.6} \textcolor{ForestGreen}{(\textsmaller{$\downarrow$19.8\%})} & \underline{95.9} \textcolor{ForestGreen}{(\textsmaller{$\downarrow$28.1\%})} & \underline{73.1} \textcolor{ForestGreen}{(\textsmaller{$\downarrow$18.0\%})} \\ \bottomrule
\end{tabular}
\caption{Out-of-domain evaluation with Llama-3.1-70B-Instruct as the solution generation model. `Trained Small Code/Math' refers to Granite-3.1-8B-Instruct which was fine-tuned on one domain in CRISP. `Err$\downarrow$' refers to the error reduction in percentage compared to the CoT baseline.}
\label{tab:out-of-distribution-results}
\end{sc}
\end{center}
\end{table*}

\subsection{Out-of-Domain Evaluation}
\label{subsec:out-of-domain-evaluation}
We hypothesize that training a model to generate high-level plans in a specific domain---such as mathematics or coding---can provide it with transferable task decomposition capabilities that enhance performance in other domains. To investigate this, We evaluated our high-level plan generation models on out-of-distribution data by fine-tuning each model on one domain and testing it on the other, i.e. the mathematics-trained model to coding tasks and vice versa.

As shown in Table~\ref{tab:out-of-distribution-results}, training on out-of-domain data still substantially improves. Moreover, training improves on out-of-domain tasks almost as much as it does on in-domain, and outperforms untrained large model. The mathematics-trained model, when applied to coding problems in the MBPP and HumanEval datasets, generated high-level plans that improved the final solution accuracy, achieving scores of 87.4 and 84.6, respectively. These results outperforms the vanilla models and are only marginally lower than those obtained by the specialized coding high-level plan generation model, which scored 87.9 and 85, respectively. Similarly, when the coding-trained model was tested on mathematical problems using the GSM8K and MATH benchmarks, it scored 95.2 and 71.9, compared to 96.9 and 73.1 achieved by the mathematics-trained model. These findings demonstrate that a model trained on one domain can indeed contribute effectively to problem-solving in another. Comparing those results with the ones in Table~\ref{tab:merged-results} shows that training on both the in-domain and the out-of-domain data further helps overall results. We take this to mean that diversity and or more high-level plan data are still valuable.

While not completely comparable, it seems that the mathematics-trained model, i.e. `Trained Small Math' in table~\ref{tab:out-of-distribution-results}, demonstrates stronger transfer performance on coding tasks compared to the coding-trained model's performance on mathematical tasks. This is evident from its error reduction, which is relatively close to that of the coding-trained model. Specifically, the average difference in error reduction between the two was 2.8\% in code generation benchmarks, compared to 8\% in math benchmarks.

This asymmetry can be attributed to the intrinsic relationship between mathematical reasoning and coding. Algorithmic programming often relies on mathematical concepts such as logic, recursion, combinatorics, probability, and number theory. 
Consequently, a model trained on mathematical problems is likely to develop robust reasoning, pattern recognition, and generalization skills—attributes that are critically important for effective coding. 
In contrast, while a model trained on coding problems may acquire knowledge of syntax and common programming patterns, it might not cultivate the deeper mathematical reasoning skills that are essential for addressing abstract or algorithmically complex tasks.

In summary, our results suggest that high-level plan generation models possess a notable degree of domain generalizability, they improve scores substantially, and the data provided and its part all contribute to performance, surpassing generation from much stronger models. The abstract reasoning and general problem-solving strategies fostered by training on mathematical problems appear to be more readily transferable to coding tasks than the reverse. This observation underscores the potential benefits of leveraging cross-domain training to enhance the versatility and effectiveness of problem-solving models. We believe this transferability also extends to other domains and topics, even those that are not inherently symbolic.

\section{Conclusions}
In this work, we introduced CRISP, a dataset for enhancing complex reasoning in large language models through structured high-level planning. 
CRISP was developed through a rigorous data generation process, leveraging existing problem-solving datasets to extract structured high-level plans, followed by an extensive filtering and validation pipeline. 
Our experiments demonstrated that fine-tuning on CRISP enables smaller models to generate higher-quality plans, outperforming much larger models across mathematical reasoning and code generation tasks. 
Additionally, our intrinsic evaluation revealed that plans generated by fine-tuned models were shorter, more concise, coherent, and complete compared to those from vanilla models. 
We also showed that high-level planning capabilities transfer effectively across domains, with fine-tuning in one domain improving performance in another. 
This highlights the generalizability of structured planning as a trainable capability that enhances reasoning efficiency across domains. 
By releasing CRISP, we aim to encourage further research into explicit planning mechanisms, structured reasoning, and their broader applications in NLP. 
Future work may explore expanding CRISP to additional domains and refining planning strategies to bridge the gap between human and machine reasoning further.

\section{Limitations}
\label{sec:limitations}
While high-level planning was shown to benefit highly from training data, and while we do release a substantial amount of data for two domains, it is likely that other domains would benefit from such datasets and would require further work to apply our methods (or new ones) to them.
Moreover, as we base our data on data that existed in other forms and for other purposes, this may not be available in other domains.

\bibliography{custom}

\appendix
\section{Appendix}
\subsection{LoRA Finetuning}
\label{appendix:lora-finetuning}
We trained Granite-3.1-8B-Instruct~\footnote{\url{https://huggingface.co/ibm-granite/granite-3.1-8b-instruct}} on CRISP with LoRA for 5 epochs, with $R=32$, $\alpha=16$, dropout ratio of 0.05\%, a learning rate of 1e-5, a cosine learning rate scheduler and a 0.0 weight decay.

\subsection{Results with Llama-3.1-8B-instruct}
\label{subsec:appendix-results-with-llama-8b}
Here are the results for another small model that we experimented with: Llama-3.1-8B0Instruct.
We did that to make sure that the results obtained with Granite-3.1-8B-Instruct are indeed representative.

\begin{table*}[]
\begin{sc}
\begin{tabular}{@{}llllll@{}}
\toprule
Planner Model & MBPP & HumanEval & GSM8K  & MATH \\
& \textbf{Pass@1\textsmaller{(Err $\downarrow$)}} & \textbf{Pass@1\textsmaller{(Err $\downarrow$)}} & \textbf{Acc.\textsmaller{(Err $\downarrow$)}} & \textbf{Acc.\textsmaller{(Err $\downarrow$)}} \\ \midrule
CoT                      & 60.6 & 72.4      & 84.3  & 51.5 \\
Vanilla Llama3.1-8B      & 60.9 \textsmaller{(\textcolor{ForestGreen}{0.8\%$\downarrow$})} & 72.6 \textsmaller{(\textcolor{ForestGreen}{0.7\%$\downarrow$})} & 84.7 \textsmaller{(\textcolor{ForestGreen}{2.5\%$\downarrow$})} & 51.9 \textsmaller{(\textcolor{ForestGreen}{0.8\%$\downarrow$})} \\
Vanilla Llama-70B        & 62.3 \textsmaller{(\textcolor{ForestGreen}{4.3\%$\downarrow$})} & 73.5 \textsmaller{(\textcolor{ForestGreen}{4.0\%$\downarrow$})} & 85.6 \textsmaller{(\textcolor{ForestGreen}{8.3\%$\downarrow$})} & 52.7 \textsmaller{(\textcolor{ForestGreen}{2.5\%$\downarrow$})} \\
Trained Llama            & 64.9 \textsmaller{(\textcolor{ForestGreen}{10.9\%$\downarrow$})} & 76.1 \textsmaller{(\textcolor{ForestGreen}{13.4\%$\downarrow$})} & 86.3 \textsmaller{(\textcolor{ForestGreen}{12.7\%$\downarrow$})} & 54.5 \textsmaller{(\textcolor{ForestGreen}{6.2\%$\downarrow$})} \\
\bottomrule
\end{tabular}
\caption{Results with Llama-3.1-8B-Instruct as a generator model with different planners, including a fine-tuned model of the aforementioned model. Notably, the trends are similar to the trends seen with Granite-3.1-8B-Instruct.}
\end{sc}
\end{table*}

\subsection{Prompt for Plan Generation in CRISP}
\label{subsec:appendix-prompt-for-plan-generation-in-crisp}
To generate plans for each domain in CRISP, we crafted a few-shot prompt for each domain.
Here is the prompt we used for the generation of plans in the Math domain of Magpie-reasoning-V1-150K.
The overall objective was to extract the logical strategy needed to solve a problem without relying on specific equations, function names, or detailed computations.

\begin{lstlisting}[language=custom]

System Message:
You are a helpful and concise assistant. You have access to:
1. A **Problem Description** that explains the problem at hand.
2. A **Detailed Solution** that fully works out how to solve the problem step-by-step.

Your goal is to produce a short, high-level plan describing how to solve the problem logically. 
This plan must not include any specific equations, function names, or detailed numerical computations. 
It should be purely indicative and helpful, outlining the logical strategy in 3-5 simple steps.

**User Message:**

**Task**
1. Read and understand the **Problem Description} below.
2. Review the **Detailed Solution** below (do not copy it).
3. From these, generate a concise, 3-5 step high-level plan that explains the logical approach needed to solve the problem.
4. The plan should be abstract and conceptual-avoid quoting or revealing detailed equations, formulas, function names, or code.
5. Focus on the reasoning steps rather than low-level implementation.

**Problem Description**
\{problem\_description\}


**Detailed Solution**
{detailed_solution}


**Formatting Requirements**
1. Your final answer should be 3-5 bullet points (or numbered steps).
2. Each bullet/step should be brief, logical, and to the point.
3. Do not include specific equations or code references.
4. Do not include extraneous commentary or repeat large sections from the solution.
5. Focus on a clear, conceptual strategy that someone could follow to solve the problem at a high level.


**Example Output Structure**
1. Identify the main elements, quantities, or variables in the problem.
2. Determine the key relationships or principles that connect these elements.
3. Outline a general strategy for combining or manipulating these elements to get closer to a solution.
4. Check or validate the approach by ensuring it aligns with the key requirements.
5. Summarize the final reasoning step or expected result in broad terms.

**Few-Shot Examples**


**Example 1**

**Problem Description**
Consider a regular octagon. How many different triangles can be formed if the octagon is placed inside a circle and we can also use the center of the circle as a vertex for the triangles?


**Detailed Solution**
Let's break it down step by step.
Step 1: Triangles with 3 vertices from the octagon (choose any 3 of 8).
Step 2: Triangles with 2 vertices from the octagon plus the center.
Then sum the totals from Step 1 and Step 2.


**High-Level Plan**
1. Recognize the two types of triangles possible: those with only octagon vertices and those that use the center as one vertex.
2. Conceptually determine how to count each type of triangle without going into specific combinations.
3. Combine the counts logically to get the total number of different triangles.


**Example 2**
**Problem Description**
Write a function that merges two sorted lists into a single sorted list, without using any built-in sorting functions. The time complexity should be \(O(n)\), where \(n\) is the total number of elements in both lists.


**Detailed Solution**
def merge\_sorted\_lists(list1, list2):
 i, j = 0, 0
 result = []
 while i < len(list1) and j < len(list2):
    if list1[i] < list2[j]:
        result.append(list1[i])
        i += 1
    else:
        result.append(list2[j])
        j += 1
 while i < len(list1):
    result.append(list1[i])
    i += 1
 while j < len(list2):
    result.append(list2[j])
    j += 1
 return result


**High-Level Plan**
1. Recognize the need to keep track of where we are in each list as we form the new list.
2. Conceptually compare the front elements from both lists to decide which goes next.
3. Continue until one list is exhausted, then add any remaining elements from the other.
4. Return the combined list as the final merged sequence.


Now, please provide your high-level plan in 3-5 steps.
\end{lstlisting}

\subsection{LLM-based Judgement Prompt}
\label{subsec:appendix-llm-as-a-judge-prompt}
We used the following prompt for judging the four attributes of generated plans with LLM-based judgement.

\begin{lstlisting}[language=custom]
You are an intelligent, knowledgeable, and impartial judge. Your task is to evaluate whether a **High-Level Plan** effectively outlines the logical steps required to address a given **Problem Description** and reach a **Detailed Solution**.

You are provided with three components:
1. **Problem Description:** {problem_description}
2. **High-Level Plan:** {high_level_plan}
3. **Detailed Solution:** {solution}

---

### **Evaluation Criteria**
Assess whether the High-Level Plan sufficiently and logically bridges the Problem Description and the Detailed Solution based on the following four criteria:

#### **1. Clarity**
- Are the steps written in a way that is easy to understand?
- Does the plan avoid ambiguity and vague language?
- Are complex ideas broken down into comprehensible components?

#### **2. Conciseness**
- Does the plan avoid unnecessary repetition or overly verbose explanations?
- Are the steps described efficiently without losing essential details?
- Is there any redundant or overly wordy content that could be simplified?

#### **3. Coherence (Logical Flow & Structure)**
- Do the steps follow a clear and logical progression from problem to solution?
- Are there any gaps, abrupt transitions, or missing links in the reasoning?
- Is the structure intuitive, making it easy to follow the problem-solving approach?

#### **4. Completeness**
- Are all necessary steps included to fully address the problem and derive the solution?
- Does the plan leave out any critical information or assume unstated knowledge?
- Are there any logical leaps where a step is missing between two points?
---

### **Output Format**
Your evaluation must be returned **as a single JSON object** containing exactly **two fields**:

1. **`explanation`**: A detailed assessment, addressing how well the plan meets each of the four criteria above. Reference specific strengths and weaknesses.
2. **`judgement`**: A string set to `"true"` if the plan **fully satisfies all four criteria**, or `"false"` otherwise.

---

### **Strict Output Requirements:**
- **Do not** include any extra keys or fields.
- **Do not** output any additional text outside the JSON structure.
- The final output must strictly match the following format:

```json
{
  "explanation": "Your detailed reasoning here.",
  "judgement": "true or false"
}
\end{lstlisting}

\subsection{Prompt for Intrinsic Evaluation}
We attach here the prompt we used to compare two plans based on clarity, conciseness, coherence, and completeness.

\begin{lstlisting}[language=custom]
You are an impartial and expert judge. Your task is to evaluate two plans that each aim to solve the same problem.
They both rely on the same problem description and reach the same final solution, but they may differ in how they
outline the logical steps to get from the problem statement to the solution.

### Your Goal
Read the problem description, the detailed solution, and both Plan A and Plan B carefully.
Then, compare and evaluate Plan A and Plan B according to four specific criteria:

1. **Clarity**
   - Are the steps written in a way that is easy to understand?
   - Does the plan avoid ambiguity and vague language?
   - Are complex ideas broken down into comprehensible components?

2. **Conciseness**
   - Does the plan avoid unnecessary repetition or overly verbose explanations?
   - Are the steps described efficiently, without omitting crucial details?
   - Is there any redundant or overly wordy content that could be simplified?

3. **Coherence (Logical Flow & Structure)**
   - Do the steps follow a clear and logical progression from problem to solution?
   - Are there any gaps, abrupt transitions, or missing links in the reasoning?
   - Is the structure intuitive and easy to follow?

4. **Completeness**
   - Are all the essential steps included to fully address the problem and derive the solution?
   - Does the plan omit any critical information or assume unstated knowledge?
   - Are there any logical leaps or missing transitions between key points?

### Inputs
**Problem Description:**
{problem_description}

**Detailed Solution:**
{solution}

**Plan A:**
{planA}

**Plan B:**
{planB}

Each plan proposes a logical sequence of steps to move from the problem description to the final solution.

### What to Do
1. Examine each plan in relation to the problem and the solution.
2. Assess Plan A and Plan B based on the four criteria: Clarity, Conciseness, Coherence, and Completeness.
3. Decide whether Plan A is superior, Plan B is superior, or they are equally good overall.

### How to Report
Provide your final output as a single JSON object in the exact format below:

{{
  "explanation": "Explain your comparison referencing each of the four criteria as needed.
                 Describe strengths, weaknesses, and the reasoning leading to your final verdict.",
  "judgement": "A or B or Same"
}}

- **explanation**: Briefly but comprehensively summarize the comparison, indicating why you believe
  Plan A or Plan B is better, or why they are the same. Please point to relevant details from each plan
  when forming your reasoning.

- **judgement**: Must be exactly one of:
  - "A" (if Plan A is judged superior overall),
  - "B" (if Plan B is judged superior overall),
  - "Same" (if they are equally good).

Ensure you base your judgment only on the given criteria and the content of the plans. Output **only** the JSON with
no additional text, headers, or explanations.

\end{lstlisting}

\end{document}